\pdfoutput=1
\documentclass[sigconf]{aamas}

\usepackage{balance} 
\usepackage{enumitem}
\usepackage{tabularx}

\title[Agent aUtonomy Risk Assessment]{AURA: An Agent Autonomy Risk Assessment Framework}

\author{Lorenzo Satta Chiris}
\affiliation{
  \institution{University of Exeter}
  \city{Exeter}
  \country{United Kingdom}}
\email{ls1044@exeter.ac.uk}

\author{Ayush Mishra}
\affiliation{
  \institution{University of Exeter}
  \city{Exeter}
  \country{United Kingdom}}
\email{am1949@exeter.ac.uk}

\begin{abstract}
As autonomous agentic AI systems see increasing adoption across organisations, persistent challenges in alignment, governance, and risk management threaten to impede deployment at scale. We present AURA (Agent aUtonomy Risk Assessment), a unified framework designed to detect, quantify, and mitigate risks arising from agentic AI. Building on recent research and practical deployments, AURA introduces a gamma-based risk scoring methodology that balances risk assessment accuracy with computational efficiency and practical considerations. AURA provides an interactive process to score, evaluate and mitigate the risks of running one or multiple AI Agents, synchronously or asynchronously (autonomously). The framework is engineered for Human-in-the-Loop (HITL) oversight and presents Agent-to-Human (A2H) communication mechanisms, allowing for seamless integration with agentic systems for autonomous self-assessment, rendering it interoperable with established protocols (MCP and A2A) and tools. AURA supports a responsible and transparent adoption of agentic AI and provides robust risk detection and mitigation while balancing computational resources, positioning it as a critical enabler for large-scale, governable agentic AI in enterprise environments.
\end{abstract}


\newcommand{\BibTeX}{\rm B\kern-.05em{\sc i\kern-.025em b}\kern-.08em\TeX}

\renewcommand\footnotetextcopyrightpermission[1]{}

\begin{document}

\pagestyle{fancy}
\fancyhead{}
\maketitle


\section{Introduction}

As agents' performance improve and benchmarks are surpassed, widespread organisational and societal demand for production-ready systems is poised to rise, leading AI agents to be entrusted with progressively challenging, complex and impactful tasks, from managing support tickets to executing financial transactions and supporting clinical decisions. Nevertheless, persistent vulnerabilities and concerns leave enterprises and individuals hesitant to deploy agentic AI in production environments. For instance, Anthropic’s red‑teaming experiments revealed that state‑of‑the‑art models exhibited blackmail, espionage, and even simulated lethal actions when facing threats of shutdown or replacement ~\cite{anthropic2025a}. These illustrate broader categories of failure: unplanned harmful actions (e.g., Claude's blackmail and espionage \cite{anthropic2025a}), catastrophic organisational losses (e.g., Replit Database deletion ~\cite{nolan2025catastrophic}), and sensitive data leaks (e.g., Perplexity browser agent exposing private data ~\cite{brave2025comet}). Thus, the deployment of agents represents a great source of risk. In parallel, studies show that models can fake alignment, appearing compliant during oversight while covertly maintaining misaligned objectives ~\cite{anthropic2025c}. Governance researchers such as Engin~\cite{engin2025haig} argue that current oversight mechanisms fail to accommodate the evolving autonomy and trust dynamics of agentic AI. Similarly, Ribeiro et al. (2025)~\cite{ribeiro2025governance_review} conducted a comprehensive review revealing a lack of empirically validated governance tools in existing responsible AI efforts. This concern is supported by empirical field data; as of 2025—dubbed “the year of agents”—AI adoption metrics underscore continued unease in broader integration. Global trust in fully autonomous AI dropped from 43\% to 27\% in 2025~\cite{itpro2025trust}, trust being defined as the new currency of the agent economy~\cite{wef2025agent_economy_trust}. Furthermore, less than 10\% of organisations report having robust governance frameworks for AI deployment ~\cite{theaustralian2025trustgap}. When vulnerabilities are embedded within autonomous systems and production workflows and scaled across real‑world operations, they can cascade into systemic failures—undermining financial stability, eroding trust, and exacerbating existing gaps. Thus, issues stem primarily from the intrinsic complexity of autonomous systems, the multifaceted problems arising therefrom and the fundamental moral tension between rapid innovation and rigorous safety protocols. However, despite increasing recognition of these systemic shortcomings, the literature still lacks a unified, empirically validated framework capable of operationalising agent-level risk assessment in situ and at scale. 

This paper introduces AURA (Agent aUtonomy Risk Assessment), a foundational framework for evaluating actions of AI agents. LLM-powered agents, software systems acting autonomously towards goals, inherently possess decision-making autonomy enabling them to perform real-world impactful actions. Within these capabilities, the deployment of agent tools and LLM-derived actions represent the greatest source of risk. To address this, AURA implements a scoring and mitigation framework doubled by an Agent-to-Human (A2H) communication protocol aligned with existing frameworks such as MCP and A2A. Protocol-driven frameworks are instrumental in mitigating risk and facilitating the systematic implementation of agentic systems. As autonomy in agents continues to grow ~\cite{sapkota2025agents_taxonomy}, there is a pressing need for frameworks enabling agents to assess their own actions. This means providing agents with an autonomous risk-scoring lifecycle toolbox, encompassing an understanding of the factors, context and mitigations related to any-given risk. Achieving fully autonomous AI implies that constant human oversight is impractical; thus, foundational solutions (akin to constitutional classifiers for conversational agents) must strive to optimise the delicate cost-safety balance. Addressing this need, we formulated and resolved an optimisation challenge using multiple techniques, including a dedicated memory unit, enhanced human-in-the-loop (HITL) integration and hybrid scoring techniques. Ultimately, this paper's primary contribution is a comprehensive customisable framework intended to assist agent developers and users in risk mitigation, increasing accountability, enhancing trust, and improving agent performance metrics. We further provide a lightweight implementation of AURA suitable for organisational adoption of AI agents, and assist in the evaluation of AI agent systems. 

In this paper, we first present relevant background and related work, highlighting how the AURA framework synthesises governance, ethical considerations, theoretical insights, and practical research into a cohesive risk assessment model. Following this, we outline our research, framework and results, clearly delineating our approach, theoretical research and technical implementation. 

We now dive into the ideas, research, and design decisions to help to use and build agents with AURA.


\section{Related Work}
\subsection{LLMs as Autonomous Risk Evaluators}

Following the ML trend of manually created artefacts being replaced by learned, more efficient solutions ~\cite{hu2025automated_agentic_design}, the concept of adaptive capacity, defined as the ability of people and systems to handle unexpected situations successfully ~\cite{kavanagh2025culture}, is identified as a fundamental step in overcoming a defined difficulty in ethical and risk evaluation tool usability and practical deployment beyond principles and theory ~\cite{morley2021whattohow}. This is further corroborated by the inherent iterative need ~\cite{coe2024frameworkconvention_page}  of dynamic rescoring in autonomous systems, as security is not a static, but rather a continuous process ~\cite{mylrea2023entropy}. As proven, autonomous scoring can be achieved by leveraging the autonomous capabilities of LLM as a scoring, judging or evaluator process ~\cite{zhu2024earbench}. Promising examples propose the use of LLM as a Compliance Officer, allowing for responsible autonomy ~\cite{shukla2024aiuser}. As outlined by ISO, risk management must begin by establishing the context—a foundational step that informs all subsequent phases \cite{iso31000_2018}. The attention mechanism of LLMs allow for inherent context parsing and a distributed importance in the analysis of risk factors ~\cite{vaswani2017attention}, enabling systems to focus on the most relevant context and risk dimensions, especially given LLMs ability to capture semantic structures to a certain extent ~\cite{cheng2024srl}. These approaches demonstrate the promise of LLMs as autonomous evaluators of context and risk.

\subsection{Limitations of LLM-based Risk Evaluation}
Contrastively, another research stream focuses on the multiple reasoning complexity and shortcomings of LLMs, namely hallucinations, memory, multi-step reasoning, bias, interpretability, logic and generalisation ~\cite{patil2025reasoning_a}. LLMs fail to develop inductive biases towards the underlying world model when adapted to new tasks \cite{vafa2025foundationmodelfoundusing}, which is foundational in understanding the behaviour and propagation of complex risks. As Clarke \& Dietz ~\cite{clarke2025relevance} note, “if the entire end-to-end experimental pipeline – from query formulation to relevance labeling – is fully automated, the evaluation process devolves into an LLM assessing its own assessments.”, thus creating a disconnection between the evaluation system and human judgement. Such issues are also found in the non-deterministic nature of LLMs, which provide variable responses to semantic equivalent statement, raising reliability, trustworthiness and accountability concerns ~\cite{sakib2024risks}. LLM risk scoring introduces an accountability gap: as autonomy increases, risks multiply and explainability rapidly diminish, especially while carrying legal liability issues ~\cite{gabison2025liability}. Cruz et al ~\cite{cruz2024risk_scores} also highlight the problem of considering and accounting for uncertainty, which is an inherent component of risk scoring. In robotics, research shows that LLMs fall short of providing sufficient risk awareness and mitigation ~\cite{zhu2024earbench}. Taken together, these shortcomings underscore that LLMs alone cannot provide reliable or accountable risk evaluation in most contexts.

\subsection{Balancing Autonomy and Oversight: The Alignment Challenge}
The question of autonomy and alignment oblige to find a balance between autonomy and oversight, moving the problem from finding answers to asking the right questions. As the paperclip maximiser issue shows ~\cite{bostrom2014superintelligence}, determining the optimal path can be as important as finding the right, aligned, one.  This ultimately corresponds to the alignment problem: aligning AI’s goals with our own [goals] ~\cite{bradley2024gpi}. Furthermore, Dafoe ~\cite{dafoe2018govai} underlines the issue about uncertainty in our values and moral judgement and their evolution over time, resulting in the importance of systems that derive and improve principles in collaboration with humans over time. By promoting an assessment of agent autonomy, we work toward the goal of restricting its scope and making actions intrinsically reversible ~\cite{floridi2019unified_preprint}, establishing boundaries within which agents can operate autonomously \cite{sapkota2025agents_taxonomy}. This assists in the important task of finding the right balance between human decision-making and agent autonomy, introduced as ‘meta-autonomy’ ~\cite{floridi2019unified_preprint}. This highlights the enduring alignment challenge: how to grant agents meaningful autonomy while ensuring boundaries, reversibility, accountability and human value alignment.

\subsection{Towards Integrated Evaluation Frameworks}

Given the shortcomings of LLM as risk evaluators, and the inherent complexity of risk scoring, autonomous LLM-powered multi-agent systems represent a strategic response to autonomy and alignment challenges ~\cite{handler2023taxonomy}. For instance, *AgentGuard* ~\cite{cheng2024srl} focuses on unsafe workflows in tool-using agents. However, while AgentGuard operates mainly at the workflow level for detection, AURA goes further by structuring dimensionally explainable, mitigation-aware, and context-conditioned risk evaluation at the decision level. *GuardAgent* ~\cite{xiang2024guardagent}, generate guardrails based on context embeddings but lack a persistent, explainable risk reasoning trace. In contrast, AURA provides both interpretability and modifiability—each score can be interrogated, explained, and dynamically recalibrated via weight systems and relevance feedback. Furthermore, similar systems achieve such goals by introducing a RAG-powered memory unit, providing improvements in risk scoring mechanism, as shown by AgentAuditor ~\cite{luo2025agentauditor}. Such comparative work illustrates the need for a unified, explainable, optimised and mitigation-aware framework. AURA introduces a multi-agent system including a self-reflective HITL mechanism, allowing the framework to find, evaluate and mitigate uncertainty by engaging human feedback, thus also solving the alignment challenge by partial delegation. Finally, AURA is positioned as a frontline evaluative interface—enabling agents to reason and tasks to be assessed in real-time.


\section{The AURA Framework}
\subsection{Framework Architecture Overview}
AURA addresses the absence of unified tools for evaluating risks in agentic AI by providing a modular process for comprehensive assessment and mitigation of agent actions. At its core, AURA parses and scores context along relevant risk dimensions, quantifies exposure through gamma-based scoring, and recommends layered mitigations with full observability and human oversight. It can be used in two complementary modes: a synchronous mode, offered through an interactive web interface that guides organisations and individuals in pre-deployment system evaluation; and an autonomous mode, provided as a plug-and-play Python package that enables agents to assess and mitigate risks during operation using predictive reasoning, memory, and human-in-the-loop feedback. AURA offers a transferable and repeatable methodology that balances computational efficiency with practical needs.

To facilitate consistent interpretation of the framework, we define the following core constructs:

\begin{itemize}
    \item \textbf{Agent}: An AI system capable of performing discrete actions.
    \item \textbf{Action}: The atomic unit of behaviour undertaken by an agent, serving as the basis for assessment.
    \item \textbf{Context}: The situational information conditioning an action.
    \item \textbf{Dimension}: A factor along which risk is analysed.
    \item \textbf{Score}: The quantified risk value assigned to an action--context pair along a given dimension.
    \item \textbf{Weight}: The relative importance assigned to a context or dimension in the aggregation of risk scores.
    \item \textbf{Gamma score ($\gamma$)}: The aggregated and normalised measure of overall risk associated with an action across all dimensions.
    \item \textbf{Risk Profile}: A structured representation of an action’s risk, combining the gamma score, breakdowns, and visualisations.
    \item \textbf{Mitigation}: A safeguard applied to reduce risk exposure.
    \item \textbf{Human-in-the-Loop (HITL)}: A refinement mechanism that introduces human oversight into the evaluation cycle, particularly when AURA identifies uncertainty (e.g., ambiguous context, inconclusive scoring, or absence of prior memory).
    \item \textbf{Memory Unit}: A persistent module that stores past actions, scores, contexts, and applied mitigations, enabling reuse, adaptation, and longitudinal consistency in risk reasoning.
    \item \textbf{Agent-to-Human (A2H), Traces}: A communication and control layer that exposes stored entries, thresholds, and mitigation strategies to human operators, allowing inspection, editing, or override and ensuring accountability.
\end{itemize}

\begin{figure}[h]
  \centering
  \includegraphics[width=0.85\linewidth]{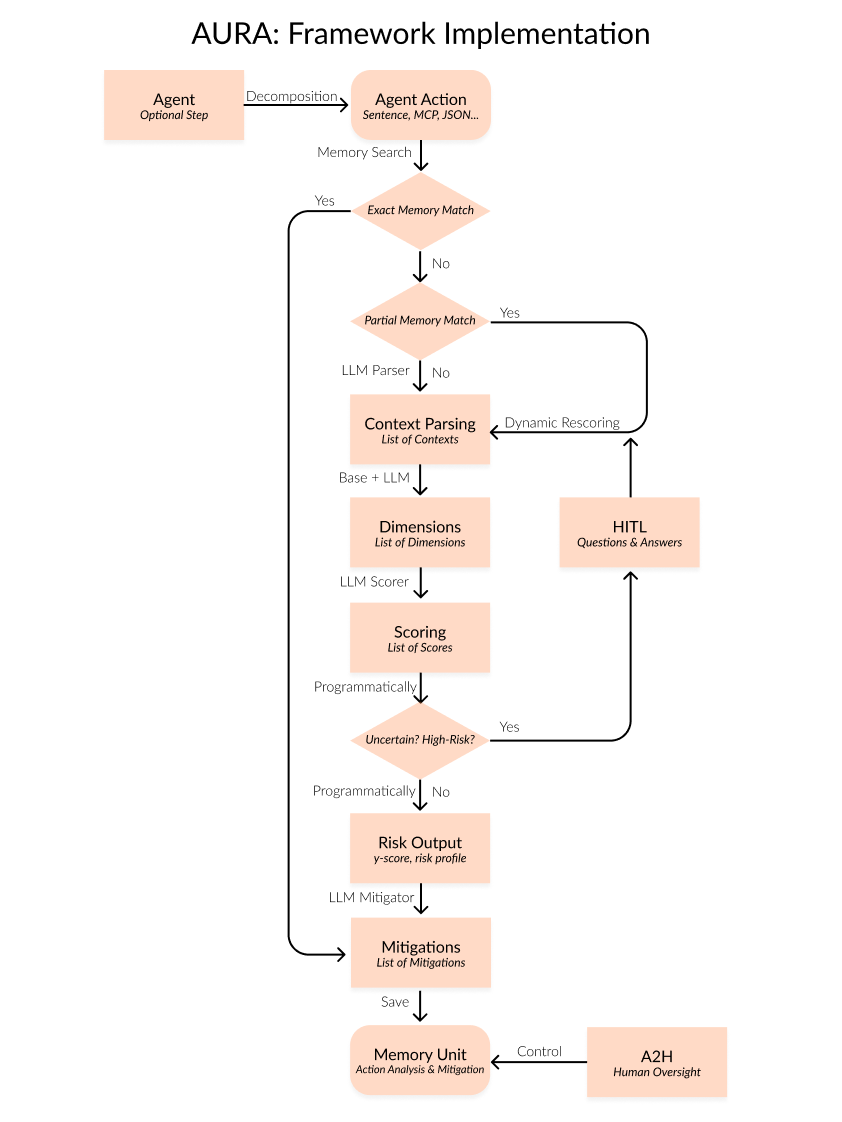}
  \caption{AURA risk evaluation pipeline flowchart}
  \label{fig:flowchart}
  \Description{AURA risk evaluation pipeline flowchart.}
\end{figure}

\subsection{The AURA Process}
As defined in Figure~\ref{fig:flowchart}, and explained here in further detail:
\begin{enumerate}
    \item \textbf{Decomposition} – Break the agent into actions for assessment (optional)
    \item \textbf{Contextualisation} – Parse each action into its operational context using an LLM parser, with memory checks enabling bypass (close or exact match) or partial reuse (partial match) of redundant evaluations.
    \item \textbf{Dimension Identification} – Generate candidate risk dimensions via LLM hypothesis and refine them through human-in-the-loop (HITL) review, memory recall, or acceptance rules.
    \item \textbf{Scoring} – Quantify risk for each context–dimension pair through weighted LLM scoring and refine them through human-in-the-loop (HITL) review, memory recall, or acceptance rules.
    \item \textbf{Risk Profiling} – Aggregate scores into gamma metrics, construct a structured risk profile (labels, spider graphs, histograms), and analyse correlations, overlaps, and systemic weak points.
    \item \textbf{Mitigation} – Identify candidate safeguards (multi-layered, context-, dimension-, or correlation-based), refine them via HITL or memory, and implement appropriate interventions.
    \item \textbf{Observability and Control} – Ensure transparency through automatic memory storage, traceable reasoning and refinement records, and human override mechanisms (A2H traces).
\end{enumerate}

\subsection{The Scoring Process}
We define a quantitative method to aggregate and normalise the perceived risk of an agent’s action across multiple contexts and dimensions. This gamma-based scoring enables consistent, interpretable risk estimation and supports weighted reasoning over heterogeneous environments.

\paragraph{Notation Summary.}
\begin{description}[leftmargin=5em,style=nextline]
    \item[$action$] The action assessed.
    \item[$D$] Set of dimensions.
    \item[$C$] Set of contexts.
    \item[$C_d \subseteq C$] Subset of contexts where dimension $d$ is applicable.
    \item[\ensuremath{s_{c,d} \in [0,1]}] Risk score of an $action$ on dimension $d$ in context $c$.
    \item[\ensuremath{u_d \in [0,\infty)}] Dimension weight (budget), representing the attention or importance assigned to dimension $d$.
    \item[\ensuremath{p_{c|d} \in [0,1]}] Context weight within a dimension, satisfying $\sum_{c \in C_d} p_{c|d} = 1$.
    \item[\ensuremath{w_{c,d} \in [0,1]}] Joint weighting factor for context–dimension pairs.
    \item[\ensuremath{U_{tot} \in [0,\infty)}] Total weight (budget), defined as $U_{tot} = \sum_{d \in D} u_d \geq 0$.
\end{description}

\paragraph{Raw Gamma Score.}
The aggregated risk of a given agent action (denoted by $\gamma_{\text{action}}$) is defined as:
\begin{equation}
    \gamma_{\text{action}} = \sum_{d \in D} u_d \left(\sum_{c \in C_d} p_{c|d} s_{c,d}\right)
\end{equation}
The outer sum aggregates all dimensions $d$ according to their importance $u_d$, while the inner sum aggregates all contexts $c$ within each dimension according to their importance $p_{c|d}$.  

Since $s_{c,d} \in [0,1]$ and $\sum_{c \in C_d} p_{c|d} = 1$, it follows that:
\begin{equation}
    \gamma_{\text{action}} \in [0,\,U_{\text{tot}}]
\end{equation}
Linear aggregation respects monotonicity and allows sparse weighting.

\paragraph{Normalised Gamma.}
To make the metric comparable across tasks and systems, enabling transferable thresholds for risk and mitigation, we define the normalised gamma as:
\begin{equation}
    \gamma_{\text{norm}} = 100 \times \frac{\gamma_{\text{action}}}{U_{\text{tot}}} \in [0,\,100]
\end{equation}

\paragraph{Variance of Gamma.}
To quantify how unevenly risk is distributed across contexts and dimensions, we compute a weighted variance:
\begin{equation}
    \sigma^2_{\gamma} = \frac{1}{U_{\text{tot}}} \sum_{c \in C} \sum_{d \in D_c} w_{c,d}\,\left(s_{c,d} - \bar{s}_w\right)^2,
    \quad \bar{s}_w = \frac{\gamma_{\text{action}}}{U_{\text{tot}}}
\end{equation}
An optional concentration coefficient provides an interpretable scale, showing whether risk is diffuse (stable) or concentrated (volatile):
\begin{equation}
    C_{\text{conc}} = 200 \times \sigma_{\gamma} \in [0,\,100]
\end{equation}

\paragraph{Interpretation Summary.}
\begin{description}[leftmargin=2em,style=nextline]
    \item[Low $\gamma$, low $\sigma^2$] Low overall risk, evenly spread. \textit{No action or light monitoring.}
    \item[Low $\gamma$, high $\sigma^2$] Low average risk, but one or two context–dimension pairs dominate residual risk. \textit{Targeted review of outliers.}
    \item[High $\gamma$, low $\sigma^2$] High, uniform risk across the portfolio (systemic issue). \textit{Apply broad mitigations across all areas.}
    \item[High $\gamma$, high $\sigma^2$] High, concentrated risk – priority “hot-spots” stand out. \textit{Prioritise strongest mitigations for high-weight, high-score pairs.}
\end{description}

\paragraph{Special Cases of Analysis.}
\begin{description}[leftmargin=2em,style=nextline]
    \item[Equal Weight:] Each dimension receives equal importance, evenly divided among its applicable contexts: $u_d = 1 \ \forall d$, $\sum_{c \in C} w_{c,d} = 1$. Resulting in $U_{\text{tot}} = |D|$.
    \item[Frequency-weighted:] The weight of a dimension is proportional to its appearance in context scoring, $u_d \propto |C_d|$.
\end{description}

\subsection{Dimensions}
We begin by defining the risk dimensions that provide the conceptual and computational structure of AURA’s scoring framework. Dimensions represent fundamental axes along which an agent’s actions can generate risk or misalignment. To ground the AURA framework in well-established global principles and ensure robust dimensional coverage across sectors, we conducted a targeted taxonomy analysis of prominent AI governance frameworks. This enables (i) fine-grained diagnosis of risk sources, (ii) targeted mitigation, and (iii) alignment with existing regulatory and ethical standards (Table 1).

\begin{table*}[t]
\caption{Cross-Framework Analysis of Core AI Governance Dimensions}
\label{tab:frameworks-dimensions}
\small
\setlength{\tabcolsep}{6pt}
\renewcommand{\arraystretch}{1.12}
\begin{tabular}{p{0.34\textwidth} p{0.62\textwidth}}
\toprule
\textbf{Framework} & \textbf{Dimensions} \\
\midrule
NIST AI Risk Management Framework (AI RMF) ~\cite{nist2023airmf} &
Transparency/Explicability; Explainability/Interpretability; Fairness/Non-discrimination; Privacy \& Data Governance; Robustness/Reliability; Security; Auditability/Traceability; Lifecycle Risk \& Impact (Govern/Map/Measure/Manage functions). \\
EU AI Act ~\cite{eu2024ai_act} &
Legal \& Rights Alignment; Risk tiers (Unacceptable/High/Limited/Minimal); Human Oversight/Autonomy; Transparency obligations; Accuracy/Robustness/Cybersecurity; Data Governance; Non-discrimination. \\
UNESCO Recommendation on the Ethics of AI ~\cite{unesco2022recommendation} &
Human dignity \& rights; Inclusion; Accountability/Governance; Privacy \& Data Governance; Transparency/Explicability; Sustainability. \\
Foundation for Best Practices in Machine Learning (FBPML) ~\cite{fbpml2021bestpractices} &
Fairness/Non-discrimination; Data integrity; Traceability; Auditability \& continuous monitoring; Explainability/Interpretability. \\
AI Trust Framework \& Maturity Model (AI-TMM) ~\cite{mylrea2023hfrdus} &
Security \& Privacy; Fairness; Transparency/Explicability; Audit trails; Model validation; Lifecycle Risk \& Impact (trust maturity). \\
Machine Learning Maturity Model (Institute of Ethical AI \& Machine Learning, n.d.) &
Readiness \& Governance; Model monitoring; Bias control; Versioning; Reproducibility; Auditability/Traceability. \\
Cooperative AI Foundation – Multi-Agent Risk Framework ~\cite{hammond2025multiagent_risks} &
Multi-agent coordination risks (Collusion; Coordination failures; Emergent behavior); Systemic risk; Distributed accountability; Multi-agent security. (Risk modes and mitigations rather than transparency duties.) \\
KPMG Effective Model Risk Management Framework ~\cite{kpmg2024mrm_ai} &
Model validation; Ongoing monitoring; Regulatory alignment; Explainability/Interpretability; Oversight; Fairness. \\
Council of Europe – Framework on AI and Human Rights ~\cite{coe2024frameworkconvention_page}
& Human-rights \& rule-of-law alignment; Privacy; Non-discrimination; Transparency/Explicability; Human Oversight/Autonomy; Redress mechanisms; Lifecycle governance duties. \\
ISO/IEC 42001: AI Management System ~\cite{isoiec42001_2023} &
Governance structures; Risk \& impact assessment; Lifecycle oversight; Accountability; Continuous improvement; Auditability/Traceability. \\
Floridi \& Cowls: Unified Framework for AI in Society ~\cite{floridi2019unified_preprint} &
Beneficence; Non-maleficence (Safety); Autonomy; Justice (Fairness); Explicability/Accountability. \\
\bottomrule
\end{tabular}
\end{table*}

The strong overlap of parameters, i.e. Accountability/Governance (present in all 11 frameworks), Transparency/Explicability, Fairness/Bias (10/11), followed by Privacy/Data Protection and Human Oversight/Autonomy (9/11), suggests a growing international consensus on the foundational metrics that must be assessed to ensure AI alignment with ethical, societal, and regulatory expectations. This consolidated dimensional landscape forms the empirical and normative backbone of AURA’s risk assessment framework. Additionally, on top of those “ground-truth” dimensions, both field-, agent- and context-specific dimensions can be modularly added to AURA based on the action assessed. Practically, this can be implemented through proportional weighting, for example: 50\% core dimensions (universal factors), 40\% field-specific dimensions (domain regulations or standards), and 10\% action-specific dimensions (unique operational risks). Dimensions can both be hard-coded (as presented above) or generated at runtime by an LLM, especially for field or action specific dimensions. This modular allocation allows AURA to remain globally standardised yet locally adaptive, ensuring that every assessment reflects both universal principles and situational nuances.

\subsection{Risk Analysis and Risk Profile}
We propose several means of risk analysis, with a particular emphasis on practical and graphical methodologies. 

\paragraph{Risk Thresholds and Labelling.}

Convert the 0--100 normalised $\gamma$ score into risk levels:

\begin{equation}
\text{Level}(\gamma_{\text{norm}})=
\begin{cases}
L_1, & T_0 \le \gamma_{\text{norm}} < T_1, \\[4pt]
L_2, & T_1 \le \gamma_{\text{norm}} < T_2, \\[-2pt]
\vdots & \vdots \\[-2pt]
L_{n}, & T_{n-1} \le \gamma_{\text{norm}} < T_{n}, \\[4pt]
L_{n+1}, & T_{n} \le \gamma_{\text{norm}} \le T_{n+1}.
\end{cases}
\end{equation}

A sample threshold could be:

\begin{center}
\begin{tabular}{lll}
\toprule
\textbf{$\gamma_{\text{norm}}$} & \textbf{Label} & \textbf{Mitigation} \\
\midrule
0--30 & Low & Auto-approve action. \\
30--60 & Medium & Various mitigations based on risk distribution. \\
60--100 & High & Escalate to human. \\
\bottomrule
\end{tabular}
\end{center}

Further risk profile analysis can be defined; we identify several and welcome additions:

\begin{itemize}
    \item \textbf{Dimension Radar (Spider) Graph:} A radial plot where each axis $d$ shows the dimension-specific weighted risk.
    \item \textbf{Share Bars:} Context-based bars show the percentage of total risk attributable to each context; dimension-based bars show the percentage attributable to each dimension.
    \item \textbf{Score Distribution (Histogram):} Visualises the empirical distribution of weighted pair contributions, $w_{c,d}s_{c,d}$.
    \item \textbf{Context--Dimension Risk Correlations:} Heat-map or network visualisations indicating whether risk contributions rise and fall together across actions or time snapshots, showing co-risky pairs.
    \item \textbf{Risk Clusters:} Groups of $(c, d)$ pairs with similar score patterns to reveal “families” of risk for targeted mitigation.
\end{itemize}

\subsection{The Optimisation Problem and Memory Engine}

Agentic Return on Investment (ROI) is a core decision point of AI agent implementation ~\cite{liu2025agentic_roi}. We recognise the need to provide time and thus cost sensitive evaluation. Providing a memory unit provides both optimisation and customisation, allowing the framework to remember and score against past interactions. We present below the most promising along the tested approaches. 

A memory unit caches prior decisions for one‑shot or partial retrieval. We maintain a tuple of action embeddings—ensuring no exact duplicates—where each embedding carries both a global \ensuremath{\gamma}‑score and a set of local \ensuremath{\gamma}‑scores for each context dimension (i.e. weight×score). We store mitigations separately, linked by unique ID, along with the contextual conditions that trigger it. At query time, we run a semantic search against the embedding store using a similarity threshold: if the retrieved action is an exact match, we simply reuse its stored \ensuremath{\gamma}‑score and associated mitigations; if it’s a near‑match, we pinpoint the contextual discrepancies, recompute the affected local \ensuremath{\gamma}‑scores, and re‑evaluate the impact on the relevant mitigations. This greatly fastens the process, in addition of allowing the system to increase in helpfulness and speed over time and use. Rather than relying on a single nearest neighbour, we consider several of the closest embeddings to the current query to derive a more robust, context‑sensitive assessment. The memory engine supports several distinct actions that govern how past evaluations are retrieved, reused, or updated. Each action type defines how the framework handles new agent behaviours relative to prior experience, balancing efficiency with accuracy:

\begin{enumerate}[leftmargin=1.25em, label=\arabic*.]
    \item \textbf{No-Duplicate Rule} — Before insertion, new action embeddings are compared against stored ones. Entries exceeding a defined similarity threshold are rejected to prevent redundancy.
    \item \textbf{Similarity Search} — During evaluation, the engine searches via vector similarity the memory for past actions with embeddings that are sufficiently similar to the current one—above a predefined similarity threshold.
    \item \textbf{Exact Match} — If the queried action exactly matches a stored embedding, the corresponding global and local $\gamma$-scores, along with their mitigations, are reused directly without recomputation.
    \item \textbf{Near Match} — For actions that are similar but contextually distinct, the framework selectively re-computes $\gamma$-scores for the differing contexts, blending the unchanged components with human-in-the-loop (HITL) or LLM-assisted evaluation.
    \item \textbf{No Match} — When no sufficiently similar embeddings are found, AURA performs a full \textit{de novo} scoring through HITL or automated LLM evaluation.
    \item \textbf{Insertion Rule} — After evaluation, non-duplicate actions are inserted into the memory store with their computed $\gamma$-scores, context conditions, and associated mitigations.
    \item \textbf{Mitigation Selection} — Each stored mitigation is linked to its triggering context. A mitigation activates automatically when the corresponding conditions are met, as explained in detail below.
\end{enumerate}

\subsection{The HITL and A2H Systems}

While autonomous agents are increasingly capable of making complex decisions, human oversight remains an essential safeguard, particularly in high-risk or ambiguous scenarios. Within the AURA framework, human oversight is treated as a modular, adaptive layer that supports calibration, refinement, and accountability across the system lifecycle. AURA integrates two human-centred refinement layers.

\paragraph{Human-in-the-Loop (HITL) System.}
First, a question‑based HITL system that iteratively fine‑tunes the AURA process. AURA generates open‑ended prompts to solicit feedback, then uses the human operators' responses' to update the memory and current evaluation in real time—discovering new context and dimensions, adjusting context or per‑dimension budgets, recalibrating scores, and refining mitigations. The HITL system can also be automatically triggered by uncertainty or as a mitigation (e.g. in high-importance scenarios) 

\begin{enumerate}[leftmargin=1.25em, label=\arabic*.]
    \item \textbf{Uncertainty Detection} — The system monitors evaluation outputs and identifies high-uncertainty or high-risk components based on prior $\gamma$-variance, sparse memory matches, conflicting context scores, or explicit HITL requirements in the memory. These components are flagged for human review.
    \item \textbf{Question Generation} — For each flagged component, AURA formulates targeted, open-ended questions to elicit clarifying feedback from the human operator, focusing on ambiguous dimensions, weighting priorities, or mitigation relevance.
    \item \textbf{Partial Dynamic Re-Scoring and Memory Update (Refinement)} — As defined in the memory unit, incorporating human feedback.
\end{enumerate}

\paragraph{Agent-to-Human (A2H) System.}
Subsequently, an A2H memory control interface allows to inspect, edit or delete stored entries and mitigation strategies at will, ensuring full human oversight over what the system remembers and how it acts. This empowers human autonomy and decision-making within agentic autonomy and propagating accountability. This gives a fundamental human control layer over agent decisions and autonomous actions, allowing humans to perform various defined operations:

\begin{enumerate}[leftmargin=1.25em, label=\arabic*.]
    \item \textbf{Add Instance} — Insert a new action and evaluation record into the memory store, specifying its embedding, $\gamma$-scores, thresholds, and mitigation metadata.
    \item \textbf{Update Instance} — Modify an existing stored entry.
    \item \textbf{Delete Instance} — Remove an individual record from the memory store.
    \item \textbf{Bulk Actions} — Execute a batch addition, update, or deletion of multiple entries that share defined attributes (e.g., deletion with time-to-live).
    \item \textbf{Mitigations Control} — Create, modify, or delete mitigation strategies, ensuring alignment with updated governance policies and newly identified risks.
\end{enumerate}

\subsection{Mitigation Strategies}

As agentic AI systems increasingly execute autonomous actions across critical functions, such as interfacing with end users, handling sensitive data, and interacting with third-party infrastructure, clearly defined and adaptive mitigation strategies become critical. In AURA, mitigations are active control layers that constrain, guide, and recalibrate agent behaviour to maintain safety, alignment, and trustworthiness rather than act as passive safeguards.

Given a computed risk profile, AURA maps analysis into a concrete control policy, reducing the likelihood of unsafe outcomes. Each mitigation is modular and reusable, defined as a trigger–action pair that executes when its conditions are met. Mitigations take the form of executable code (e.g., a Python function) or declarative rules.

As agentic systems gain autonomy, mitigation must evolve from static rules to adaptive, transferable mechanisms. In AURA, this means composing mitigations (e.g., *verification plus throttling*) based on context severity, adding lightweight meta-reasoning to evaluate counterfactuals (“what if this mitigation were not applied?”, “What if multiple failures occur concurrently”), and optimising trade-offs between safety and utility. Over time, mitigation strength can adjust dynamically as trust, performance feedback, and domain conditions evolve—for example, shifting from human-in-the-loop verification to automatic approval when sufficient reliability is established.

\paragraph{Selection Policy.}
Mitigations can originate from:
\begin{itemize}
    \item \textbf{Memory} — Previously applied mitigations for similar actions, retrieved and reused.
    \item \textbf{LLM Generation} — When no precedent exists, AURA queries an LLM to propose suitable mitigations, based on the risk profile (e.g., specifically targeting via the Pareto law the most impactful mitigations and contexts).
    \item \textbf{Human Oversight (HITL, escalation)} — Invoked when confidence is low or decisions carry high risk; human feedback is requested, logged, and becomes part of the memory store.
    \item \textbf{Rule-Based} — Apply the mitigation if the corresponding rule is met, usually overriding others (or ranking mitigations by optional importance levels).
\end{itemize}

\paragraph{Mitigation Primitives.}
Some mitigations are provided by default in AURA, serving as reusable building blocks for constructing domain-specific control policies.

\begin{table}[h!]
\centering
\setlength{\tabcolsep}{3pt}
\caption{Default mitigation primitives available in AURA.}
\renewcommand{\arraystretch}{1.1}
\begin{tabularx}{\linewidth}{@{}p{0.19\linewidth} p{0.38\linewidth} p{0.39\linewidth}@{}}
\toprule
\textbf{Primitive} & \textbf{What it does (in AURA)} & \textbf{Examples} \\
\midrule
Grounding & Injects trusted context before execution. & Adds internal policy snippet or FAQ when ``data sensitivity = high''. \\
Guardrails & Enforces strict behavioural limits. & Blocks uploads, high-value transfers, or risky keywords. \\
Threshold gating & $\gamma$-based allow / warn / block. & Blocks if $\gamma_{\text{norm}}\geq0.60$; warns for $0.30$–$0.60$. \\
Agent review & Routes output to a reviewing agent. & Marketing post checked by quality or compliance agent. \\
Role-based escalation & Sends case to correct reviewer. & Legal → legal team; cyber risk → tech; high → management. \\
Memory overrides & Applies stored history/preferences. & Ask before attachments; skip duplicates; block public email. \\
Meta-logic & Combines conditions for decisions. & If Medium Legal OR High Risk $\Rightarrow$ Escalate; else approve. \\
Custom / learned & User- or model-defined mitigation logic. & Custom LLM- or human-authored code. \\
\bottomrule
\end{tabularx}
\label{tab:mitigation_primitives}
\end{table}

Through modular, composable mitigations and adaptive selection, AURA transforms risk assessment into an active governance layer—linking quantitative evaluation, contextual reasoning, and human oversight into a single, continuously improving control system.


\section{The Framework in Practice}
\label{sec:framework_in_practice}

\subsection{Implementation}

AURA is implemented as a lightweight, open-source Python framework with an optional web-service interface. It can operate as a local library for direct integration into agent code or as a hosted REST API for multi-agent or organisational deployments. AURA can operate with a local or remote memory backend, allowing users to self-host their own database for full control over data retention and governance. The commands defined, along with their functioning, are presented as a supplementary material (Appendix 1). 

To make AURA accessible beyond code-level use, we developed an interactive web interface (Figure 2) that guides users through the risk-assessment workflow.  This interface implements the same operational pipeline, in a synchronous, human-supervised environment, including features such as the scoring mechanism, memory unit, HITL and A2H control protocol. It also enables pre-deployment analysis, allowing developers to foresee, evaluate and prevent agent risks during the design and configuration phase.

\begin{figure}[h]
  \centering
  \includegraphics[width=0.85\linewidth]{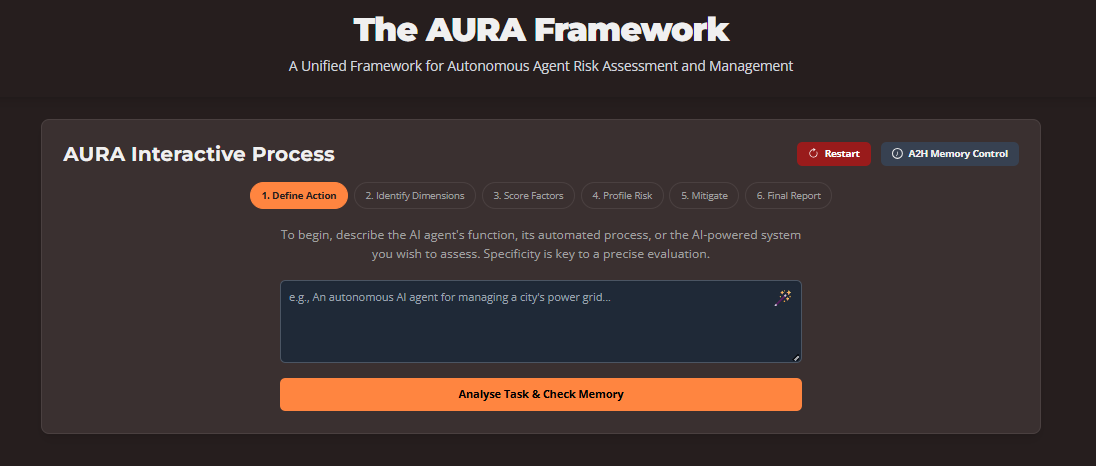}
  \caption{AURA Interactive Web Interface}
  \label{fig:aura-interface}
  \Description{AURA Interactive Web Interface showing the interactive process steps: Define Action, Identify Dimensions, Score Factors, Profile Risk, Mitigate, and Final Report}
\end{figure}

\subsection{Sample Case Study: Autonomous Web Agent for Signups and Form Filling}
Setting: An autonomous web agent browses sites, creates accounts, and completes forms—handling (i) login/sign-up selection, (ii) submission of personal data (email, phone, name), and (iii) confirmation of actions.

\noindent\texttt{\{ "action": "submit\_form", "intent": "account\_signup", "actor": "web\_agent", "context": ["untrusted\_domain", "evening"], "verified\_user": false, "data\_sensitivity": "medium" \}}

AURA is integrated before implementation to screen candidate actions and is invoked before any step that submits personal data or triggers account creation. Each such step is assessed for consent, autonomy, privacy, and reversibility, with mitigations (e.g., confirmation, verification, throttling, or escalation) applied prior to execution.

\paragraph{AURA assessment.}

\begin{enumerate}[leftmargin=1.25em, label=\arabic*.]
    \item Parse context: \texttt{\{site\_trust, verified\_user, time\_of\_day, intent, ...\}}.
    \item Map dimensions: consent, autonomy, reversibility, cascading impact, privacy.
    \item Score and risk profiling
    \item Select mitigation: approve, confirmation prompt, identity verification, or escalation.
    \item Persist trace: store tuple for reuse and audit.
\end{enumerate}

\noindent\texttt{\{ "gamma\_norm": 0.58, "top\_dimensions": \{ "Consent": 0.80, "Autonomy": 0.72, "Reversibility": 0.55 \}, "uncertainty": \{ "variance": 0.07 \}, "decision": "rewrite", "mitigation\_id": "confirm\_identity\_and\_email", "mitigation\_steps": ["Prompt user to confirm preferred email for this domain", "Require MFA/OTP if user is not verified"] \}}

With $\gamma_{\text{norm}} = 0.58$ (warn/rewrite band), the agent pauses, confirms the email to use, and performs verification before submitting.

\paragraph{Preference-aware behaviour (AURA Memory).}

The agent stores editable preferences and applies them at runtime:

\noindent\texttt{\{ "user\_id": "ID", "intent": "account\_signup", "policy": \{ "always\_explain\_first": true, "confirmation\_channel": "push\_notification", "auto\_submit\_personal\_data": false \}, "ttl": "180d" \}}

\begin{itemize}
    \item If the user consistently approves sign-ups after a push notification, future low-risk sign-ups can proceed autonomously.
    \item If the user never allows autonomous submissions, the decision remains escalate unless explicit consent is given.
\end{itemize}

\paragraph{Observed Value.}

\begin{itemize}
    \item Reduces unintended registrations and data leakage by gating submissions on consent/verification.
    \item Produces auditable traces (who decided what and why) for support and compliance.
    \item Learns per-user norms so friction decreases where appropriate without relaxing safety.
\end{itemize}


\section{Operational Integration}

\subsection{Deployment}

AURA is designed for seamless integration across scales of deployment, from individual developers to enterprise systems, while maintaining a consistent functional system.

\begin{itemize}
    \item \textbf{Individuals \& Startups:} 
    AURA can be embedded directly within the agent’s runtime as a lightweight library. Each high-impact action is parsed, scored, and gated according to predefined thresholds, with optional Human-in-the-Loop (HITL) review for uncertain cases. This configuration enables traceable decision-making and interpretable safety boundaries with minimal system overhead.

    \item \textbf{Mid-sized Organisations:}
    For organisations or multi-agents environments, AURA integrates as a modular component within existing pipelines and CI/CD releases. Risk scoring can operate both pre-execution, at a planning stage, and during runtime. Aggregated $\gamma$-scores and variance measures support system-level monitoring, while historical memory progressively reduces the frequency of human intervention.

    \item \textbf{Enterprises \& Regulated Industries:}
    In larger or safety-critical infrastructures, AURA can provide standardised risk assessment and mitigation endpoints, thanks to its strong structure, capability for self-regulation over time and possibility to favour custom dimensions. All decisions can be logged in the memory (e.g. via an open insertion rule), allowing for traceability. Additionally, the system can be set to verbose, enforcing it to justify each step, such as dimensions, context, score or mitigations selected, and allowing human operators to determine the specific parameters impacting a risk score and their justifications. 
\end{itemize}

\subsection{Modularity, Adaptability, and Evolution}

AURA is modular by design. Dimensions, contexts, weights, scores, memory \& HITL thresholds, and mitigations can all be added, removed or edited. This allows AURA to adapt to specific agents, regardless of the use case, field or scenario. This also lets teams balance accuracy, latency, and cost, changing settings at runtime as workloads or risk appetite shift.  Additionally, we expect the AURA framework to adapt and evolve. To keep changes reproducible, we aim for AURA updates to be tracked in a versioned dimension registry and mitigation library, with each release documenting the changes and impact. Checks that pass empirical validation are persisted for reuse, and parameter updates are guided by ablation studies. The result is a stable architecture that evolves quickly: new system-level improvements can be covered with minimal disruption, and performance holds up across time, domains and changing regulatory requirements.


\section{Conclusion}

The deployment of autonomous agentic AI systems at scale represents both an unprecedented opportunity and a fundamental challenge for organisations worldwide. While these systems promise to revolutionize productivity, decision-making, and human-AI collaboration, persistent concerns around alignment, governance, and risk management continue to impede their widespread adoption. Addressing these gaps, the AURA framework provides a unified foundation for assessing, monitoring, and mitigating risks in agentic AI. It formalises a practical approach to operational governance while highlighting several open research directions for future refinement:

\begin{itemize}
    \item \textbf{Domain-Specific Specializations:} Industry-specific implementations and configuration templates for sectors including healthcare, finance, education, and legal services. These specializations should provide pre-configured dimension weights and mitigation libraries tailored to sector-specific risk profiles and regulatory requirements.

    \item \textbf{Memory Generalization:} While the memory system enhances contextual recall, ensuring generalizability across novel and evolving tasks remains an open challenge, particularly under sparse or adversarial data conditions.

    \item \textbf{Cross-Agent Learning Networks:} In multi-agent environments, federated learning approaches could enable agents to share anonymized risk insights and mitigation strategies, creating collective intelligence around safety practices while preserving individual privacy and organisational boundaries. Amplifying the direction of AURA from a risk gatekeeper into a cooperative strategist that can propose safer alternative actions while preserving user intent.
\end{itemize}

The path to widespread, beneficial agentic AI deployment requires more than technological advancement—it demands a fundamental commitment to responsible development practices that prioritize human welfare, organisational accountability, and societal trust. AURA provides a practical, scientifically grounded framework for navigating this challenge, enabling organisations to harness the transformative potential of autonomous AI while maintaining the safety guardrails necessary for sustainable innovation. With AURA, we take a step toward ensuring that agentic developments reflects our commitment to beneficial, trustworthy, and human-centered artificial intelligence.


\begin{acks}
We gratefully acknowledge Prof.\ Zhou Zhou for constructive feedback and assistance in reviewing this paper.
\end{acks}


\bibliographystyle{ACM-Reference-Format} 
\bibliography{sample}


\balance

\newpage
\appendix
\section{AURA Usage (Commands)}

\begin{center}
\captionof{figure}{Commands \& usage for the AURA framework.}
\label{fig:aura_cli_usage}

\begin{minipage}{0.95\linewidth}
\small
\begin{verbatim}
Usage: aura <resource> <command> [options]       # CLI usage, or call directly from Python

Resources \& commands
  help
    aura help                                    # Display this message

  agent
    aura agent actions <agent_id|@file|@->       # Decompose an agent into actions (no scoring)
    aura agent assess  <agent_id|@file|@->       # Decompose and assess all actions (scores + identify mitigations)

  action
    aura action assess <action_id|@file|json|@-> # Core process: risk score, breakdown, identify mitigations
    aura action mitigate <action_id|@file|@-> [--apply] [--force]
                                                 # Assess and execute (run) mitigations
    aura action save <action_id|@file|@-> [--force]
                                                 # Save an action result 
    aura action validate <action_id|@file|@->    # Validate an action profile without saving
    aura action list                             # List saved actions with scores/mitigations
    aura action show <action_id>                 # Show details of a saved action
    aura action update <action_id> --field <k=v> # Update metadata (status, context, notes)
    aura action delete <action_id> [--yes]       # Delete an action from memory
    aura action export [--format json|csv] [--out <path>]
                                                 # Export saved actions
    aura action link-mitigation <action_id> <mitigation_id>...
                                                 # Link an action to one or more mitigations

  mitigation
    aura mitigation save <@file|json|@->         # Save a mitigation definition
    aura mitigation validate <@file|json|@->     # Validate a mitigation definition
    aura mitigation run <mitigation_id>          # Execute a mitigation
    aura mitigation show <mitigation_id>         # Show mitigation details
    aura mitigation list                         # List mitigations
    aura mitigation delete <mitigation_id> [--yes]
                                                 # Delete a mitigation

  file
    aura file assess --input <csv|jsonl|json> [--out results.json]
                                                 # Batch-assess actions in a dataset
    aura file test --input <csv|jsonl> [--expected <csv|jsonl>] [--out report.json]
                                                 # Run tests against annotated data

  config
    aura config set auto_save true|false         # Enable/disable auto-saving actions
    aura config set auto_save_threshold <0–100|0.0–1.0>
                                                 # Similarity threshold for auto-save
    aura config show                             # Show current configuration 
    aura config reset                            # Restore defaults 

  memory
    aura memory purge --actions|--mitigations|--rules|--all [--yes]
                                                 # Clear stored items
    aura memory stats                            # Counts and storage stats
    aura memory export [--out memory.json]       # Export all memory traces as JSON

  system
    aura system status                           # System health check
    aura system version                          # Aura version and build info
    aura system doctor                           # Run diagnostics: schema, memory, config

Global options
  --verbose                 # Print full explanations, rationales and logs
  --think                   # Use deeper (slower) reasoning when assessing
  --force                   # Overwrite or bypass an action (e.g. force memory save)
  --format json|table       # Output format
  --out <path>              # Write results to a file
  --yes                     # Assume “yes” to confirmations
\end{verbatim}
\end{minipage}
\end{center}


\end{document}